\documentclass[sigconf,nonacm]{acmart}

\usepackage[algoruled,vlined,linesnumbered]{algorithm2e}
\usepackage[textsize=small]{todonotes}

\SetAlgoSkip{SkipBeforeAndAfter}

\SetCommentSty{mycommfont}
\SetKw{Continue}{continue}

\usepackage{xspace}
\usepackage{comment}
\usepackage{graphicx}
\usepackage{xcolor}
\definecolor{gg}{RGB}{0, 155, 85}
\definecolor{primarycolor}{RGB}{33,49,77}   % dark blue
\definecolor{angrycolor}{RGB}{210,73,42}    % orange
\usepackage{amsmath} 
\usepackage{float}

\usepackage{tikz}
\usetikzlibrary{calc,fit}

\newcommand{\rrts}{\abbr{rrt*}}
\newcommand{\sst}{\abbr{sst}}
\newcommand{\stl}{\abbr{stl}}
\newcommand{\abbr}[1]{\textsc{\MakeLowercase{#1}}\xspace}

\SetKwInput{KwInput}{Input}                % Set the Input
\SetKwInput{KwOutput}{Output}              % set the Output

\definecolor{darkgreen}{RGB}{0, 128, 0}

\AtBeginDocument{%
  }

\begin{document}

%%
%% The "title" command has an optional parameter,
%% allowing the author to define a "short title" to be used in page headers.
\title{Multi-layer Motion Planning with Kinodynamic and Spatio-Temporal Constraints}

% \titlenote{Equal Contribution}
% \titlenote{Code: \url{https://github.com/elpis-lab/LG-SST-STL}}
\thanks{* Equal contribution.}
\thanks{Code: \url{https://github.com/elpis-lab/LG-SST-STL}}
  % \titlenote{*Equal contribution. \\
  % Code: \url{https://github.com/elpis-lab/LG-SST-STL}}}

%%
%% The "author" command and its associated commands are used to define
%% the authors and their affiliations.
%% Of note is the shared affiliation of the first two authors, and the
%% "authornote" and "authornotemark" commands
%% used to denote shared contribution to the research.
\author{Jeel Chatrola*}
\email{jchatrola@wpi.edu}
\orcid{0009-0008-1458-332X}
\affiliation{%
  \institution{Worcester Polytechnic Institute}
  \city{Worcester}
  \state{MA}
  \country{USA}
}

\author{Abhiroop Ajith*}
\email{aajith@wpi.edu}
\orcid{0009-0002-3986-7122}
\affiliation{%
  \institution{Worcester Polytechnic Institute}
  \city{Worcester}
  \state{MA}
  \country{USA}
}

\author{Kevin Leahy}
\email{kleahy@wpi.edu}
\orcid{0000-0001-5894-7190}
\affiliation{%
  \institution{Worcester Polytechnic Institute}
  \city{Worcester}
  \state{MA}
  \country{USA} 
}   

\author{Constantinos Chamzas}
\email{cchamzas@wpi.edu}
\orcid{0000-0001-5830-5542}
\affiliation{%
  \institution{Worcester Polytechnic Institute}
  \city{Worcester}
  \state{MA}
  \country{USA}
}
%%
%% By default, the full list of authors will be used in the page
%% headers. Often, this list is too long, and will overlap
%% other information printed in the page headers. This command allows
%% the author to define a more concise list
%% of authors' names for this purpose.
\renewcommand{\shortauthors}{Chatrola et al.}

%%
%% The abstract is a short summary of the work to be presented in the
%% article.
\begin{abstract}
% \the\columnwidth
We propose a novel, multi-layered planning approach for computing paths that satisfy both kinodynamic and spatiotemporal constraints. Our three-part framework first establishes potential sequences to meet spatial constraints, using them to calculate a geometric lead path. This path then guides an asymptotically optimal sampling-based kinodynamic planner, which minimizes an STL-robustness cost to jointly satisfy spatiotemporal and kinodynamic constraints. In our experiments, we test our method with a velocity-controlled Ackerman-car model and demonstrate significant efficiency gains compared to prior art. Additionally, our method is able to generate complex path maneuvers, such as crossovers, something that previous methods had not demonstrated. 
\end{abstract}

\received{14 November 2024}
\received[revised]{10 March 2025}
% \received[accepted]{5 June 2009}

%%
%% This command processes the author affiliation and title
%% information and builds the first part of the formatted document.
\maketitle

% \todo{Using uniform equation numbering throughout the paper}
\section{Introduction}
%Kinodynamic-based sampling-based planning and, strong algorithms  

%Sampling-based planners \cite{orthey2024-review-sampling} have become foundational in robotic motion planning, including applications where kinodynamic constraints are crucial. Recent kinodynamic planners have shown both enhanced efficiency \cite{ortiz2024idb} and asymptotically optimal performance relative to specific cost functions \cite{shome2021edges, SST}. Remarkably, these planners are capable of computing kinodynamically feasible, optimal paths without relying on a steering function, which is typically required by geometric asymptotically optimal planners like \rrts \cite{karaman2011}.

Motion planning is a core problem in robotics spanning applications from autonomous cars to long-horizon manipulation. 

Sampling-based planners \cite{orthey2024-review-sampling} have shown great promise in efficiently  computing motion plans including scenarios where kinodynamic constraints must be considered. Recent kinodynamic planners have shown both enhanced efficiency \cite{ortiz2024idb} and asymptotically optimal (\textsc{ao}) convergence for a given cost \cite{gammell2021asymptotically} enabling the efficient computation of trajectories for non-holonomic robots such as acceleration-bounded vehicles. 

\begin{figure}[H]
    \centering
    \includegraphics[width=0.85\linewidth]{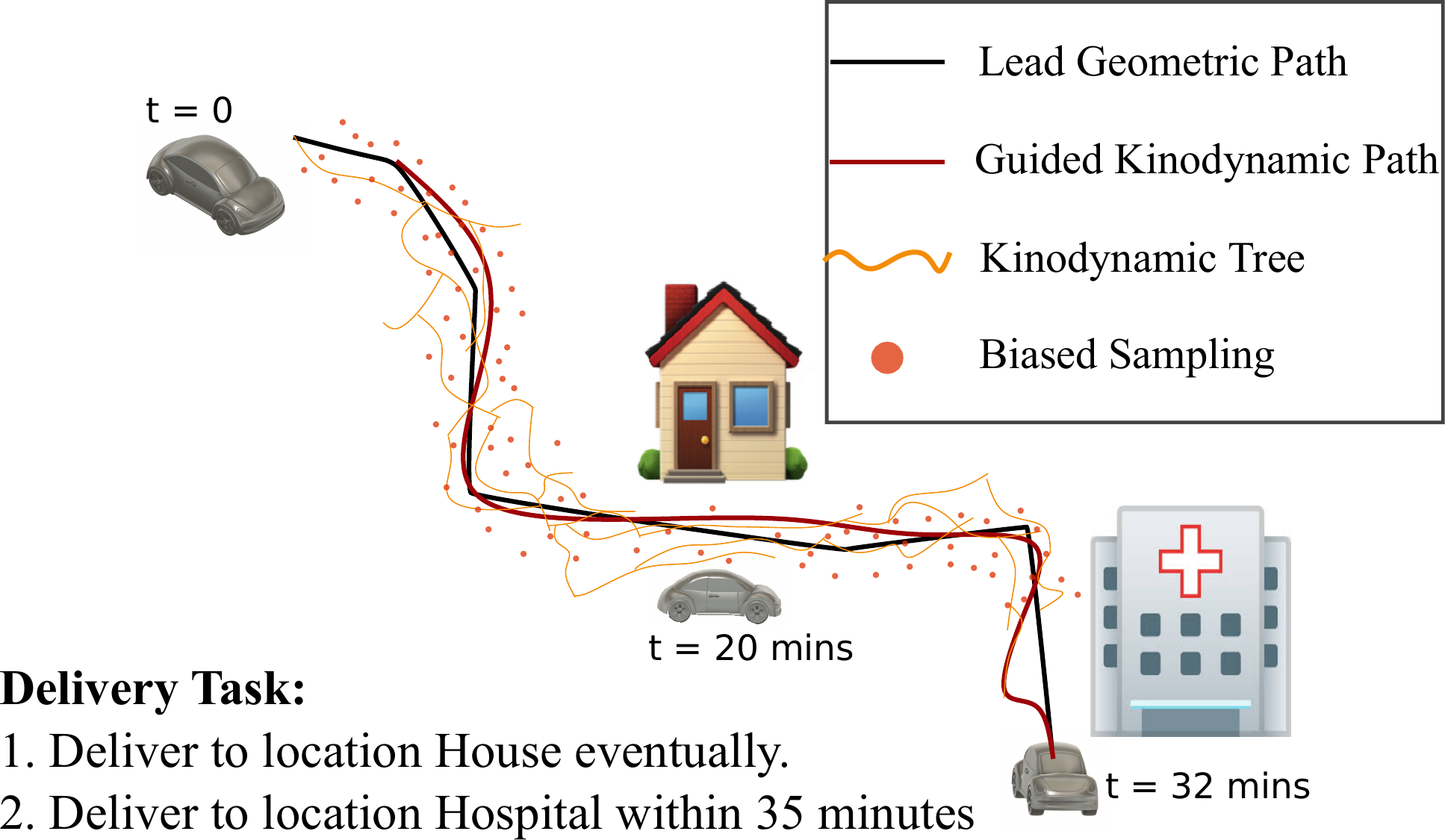}
    \caption{A delivery robot tasked with reaching different regions in a time sensitive-manner.}
    \label{fig:intro_image}
\end{figure}

However, as robotics becomes more ubiquitous and tasks grow in complexity, a single motion plan often fails to satisfy all task-specific requirements. Increasingly, robotic tasks require additional constraints to be met. For example, a delivery robot might need to visit several different regions in a time-sensitive manner as shown in \autoref{fig:intro_image}. To efficiently encode these complex mission objectives, expressive and precise logic-based tools have been used to describe the desired behavior of the system. Significant research has focused on integrating these logic-based methods with motion planning to tackle complex tasks. For instance, Linear Temporal Logic (\textsc{ltl}) \cite{LTL_paper} has been widely used to encode sequential and safety requirements \cite{plaku2012planning}, while Planning Domain Definition Language (\textsc{pddl}) is widely used for task and motion planning tasks in manipulation~\cite{garrett2021integrated}.  
% \todo{refer to this figure int the introduction}
%When complex continuous spatio-temporal constraints are present, for example, in healthcare delivery, a robot may need to visit several locations in a time-sensitive manner, as shown in \autoref{fig:example}.
Signal Temporal Logic (\stl)~\cite{STL} has found widespread applicability, as it can quantitatively monitor the satisfaction of  spatiotemporal requirements for system behavior across hybrid (discrete and continuous) domains.
Typically, \stl problems are formulated as Mixed Integer Linear Programs~(\textsc{milp})~\cite{belta2019formal,raman2014model}, which don't naturally handle kinodynamic planning problems.
Thus, researchers~\cite{scLTL-RRT*, RRT*-STL, Explo-STL} have started investigating encoding \stl robustness as part of a cost function within asymptotically optimal planners, such as \rrts\cite{karaman2011}. However, this only applies to simple kinodynamic models with easy to compute steering functions. Parallel work~\cite{ho2022automatonguidedcontrolsynthesissignal} drives the search with a time-partitioned STL automaton instead of a single robustness cost , but their method still requires a steering function—limiting applicability to complex dynamics.

%(e.g., solving the two-point boundary problem (\textsc{2bvp}), as required by the rewiring operation of \rrts).
%Additionally, as the complexity of the specification increases, the scalability of these methods becomes limited as a single cost function struggles to produce a satisfactory trajectory.

To jointly address the challenges of spatiotemporal and kinodynamic constraints, we propose a novel multi-layer framework based on Stable Sparse Tree (\sst)\cite{SST} that maintains probabilistic completeness and \textsc{ao} guarantees without requiring steering functions. Additionally, we introduce an efficient encoding scheme for the \stl formulas that enables robots to handle complex spatio-temporal constraints using geometric leads. Our approach employs a decomposition method by first generating a sequence of potential spatial regions that need to be visited. Biased sampling is used to guide exploration through these regions, providing improved efficiency, while \stl robustness is used as the cost of the \sst planner.

\section{Related Work}

%As highlighted in the introduction, kinodynamic motion planning is challenging due to the complexities of differential constraints and high-dimensional state and control spaces \cite{RKP2001}. To make this computational problem worse most of these planners rely on \textsc{2bvp} to connect to points in the state-space to find optimal control sequence ~\cite{lavalle2006planning}. 
Kinodynamic motion planning is challenging due to the complexities of differential constraints and high-dimensional state and control spaces. Finding an optimal control sequence  to connect to points in the state-space is known as the 2-point-boundary problem (\textsc{2bvp}), which is impractical to compute multiple times~\cite{lavalle2006planning}. Furthermore, the high dimensionality of the state and control spaces poses significant computational challenges, making it hard to efficiently explore and optimize solutions \cite{RKP2001}.

Over the years, several sampling-based planners \cite{orthey2024-review-sampling} tailored towards kinodynamic-based systems have been proposed that do not require solving \textsc{2bvp} problem. Examples include the original kinodynamic-RRT \cite{RKP2001}, and later layer-based improvements such as \textsc{kpiece} \cite{KPIECE} and \textsc{syclops} \cite{Syclops}. Additionally, \textsc{ao} kinodynamic planners were developed that can optimize a given cost function such as edge-bundling planners \cite{shome2021edges} and the \sst planner. However, these planners can only be used to solve a single motion plan,(e.g, reaching a single goal region) which does not suffice for more complex multi-goal missions such as in our setting.  

%When kinodynamic constraints are in place, typical \textsc{ao}, planners such as \rrts need to solve \textsc{2bvp} problem to perform the rewiring step,  thus are only applicable to systems with simple dynamics e.g., they have access to a steering function.  

To address this,  researchers have leveraged practical tools and extensive literature from formal methods to augment motion planners to synthesize controllers for temporal logic specifications \cite{kress2018synthesis}. % such as STL \cite{scLTL-RRT*}, LTL \cite{LTL_OMPL}, scLTL \cite{scLTL-RRT*}, MITL \cite{MTL}, and others \cite{Temporal-logic-basedreactive}. The various temporal logics cater to diverse applications, possessing unique, expressive capabilities that make them suited for specific tasks as seen in \cite{Vehicle_MTL, vasile2014, Explo-STL}. 
Methods most similar to our approach, which address spatio-temporal constraints, typically utilize \stl formulae to specify a given task. These methods modify the cost function of the base planner to maximize the robustness of the \stl formula, \cite{scLTL-RRT*, RT-RRT*-STL, RRT*-STL}. However, the rely on geometric \textsc{ao} planners, which restrict their applicability to systems with simple dynamics. Furthermore, as \stl formulae become more complex, encoding them into a single cost function becomes increasingly computationally expensive. In contrast, our work leverages kinodynamic \textsc{ao} planners that do not require steering functions. Additionally, we propose a simple yet complete method to reduce the complexity of the optimized robustness function, enhancing computational efficiency without sacrificing completeness.

\section{Preliminaries} 
% \todo{not sure why n is defined as the number of rows and m is defined as the number of columns. Usually, it is the other way around}
Let $\mathbb{R}, \mathbb{R}_{\geq 0}, \mathbb{N}$ denote the set of real, non-negative real, and natural numbers, respectively. We denote by $\mathbb{R}^n$ an $n$-dimensional Euclidean space and by $\mathbb{R}^{n \times m}$ a space of real matrices with $n$ rows and $m$ columns.
We use time intervals in the form $[a, b], a \leq b$. Further, we denote $t+[a, b]$ by $[t + a, t + b]$. 

Let $(M, d)$ be a compact metric space with $M \subset \mathbb{R}^n, n \geq$ 1 , and $\mathcal{S}=\left\{s: \mathbb{R}_{\geq 0} \rightarrow M\right\}$ the set of all infinite-time signals in $M$. The components of a signal $s \in \mathcal{S}$ are denoted by $s_i, i \in$ $\{1, \ldots, n\}$. The set of all linear functions over $\mathbb{R}^n$ is denoted by $\mathcal{F}=\left\{\pi: \mathbb{R}^n \rightarrow \mathbb{R}\right\}$.

% Let $\delta_\tau: \mathcal{S} \rightarrow \mathcal{S}$ be the time-shift operator acting on signals in $\mathcal{S}$ with $\tau \geq 0$, i.e., $\delta_\tau s(t)=$ $s(t+\tau)$ for all $t \geq 0$.

The syntax of STL is defined as follows \cite{STL}:
\begin{equation}
\phi::=\top \medspace|\medspace p_{\pi(x) \sim \mu} \medspace|\medspace
    \neg \phi\medspace|\medspace \phi_1 \wedge \phi_2\medspace|\medspace
    \phi_1 \mathcal{U}_{[a, b]} \phi_2,
\end{equation}
where $T$ is the Boolean true constant; $p_{\pi(x) \sim \mu}$ is a predicate over $\mathbb{R}^n$ parameterized by $\pi \in \mathcal{F}, \mu \in \mathbb{R}$ and an order relation $\sim \in\{\geq,>, \leq,<\}$ of the form $p_{\pi(x) \sim \mu}=\pi(x) \sim \mu ; \neg$ and $\wedge$ are the Boolean operators for negation and conjunction, respectively; and $\mathcal{U}_{[a, b]}$ is the bounded temporal operator \emph{until}. The Boolean semantics of STL is defined over signals in $\mathcal{S}$ recursively as follows \cite{STL}:
\begin{equation}
\begin{array}{ll}
(s,t) \models \top & \Leftrightarrow \top \\
(s,t) \models p_{\pi(x) \geq \mu} & \Leftrightarrow \pi(s(0)) \geq \mu \\
(s,t) \models p_{\pi(x) \leq \mu} & \Leftrightarrow \pi(s(0)) \leq \mu \\
(s,t) \models \neg \phi & \Leftrightarrow \neg((s,t) \models \phi) \\
(s,t) \models\left(\phi_1 \wedge \phi_2\right) & \Leftrightarrow\left(s \models \phi_1\right) \wedge\left((s,t) \models \phi_2\right) \\
(s,t) \models(\phi_1 \mathcal{U}_{[a, b]} \phi_2) & \Leftrightarrow \exists t' \in [a,b] s.t. s(t')\models \phi_2  \wedge \\& \forall t'' \in [0,t') s(t'')\models \phi_1\:,
\end{array}
\end{equation}
A signal $s \in \mathcal{S}$ is said to satisfy an STL formula $\phi$ if and only if $s(0) \models \phi$. The Boolean value false $\perp \equiv \neg \top$ and additional operations (i.e., disjunction, implication, and equivalence) are defined in the usual way. Also, the temporal operators eventually and globally are defined as $F_{[a, b]} \phi \equiv \top \mathcal{U}_{[a, b]} \phi$ and $G_{[a, b]} \phi \equiv \neg F_{[a, b]} \neg \phi$, respectively. In addition to Boolean semantics, STL admits quantitative semantics \cite{fainekos2009robustness,donze2010robust}, which are formalized by the notion of the robustness degree. The robustness degree of a signal $s \in \mathcal{S}$ with respect to an STL formula $\phi$ is a functional $\rho(s, \phi)$ recursively defined as:
\begin{equation}
\begin{array}{ll}
% \rho\left(s, p_{\pi(x) \sim \mu},t\right) & =(-1)^L(\pi(s(0))-\mu) \\
\rho(s, p_{\pi(x) \geq \mu},t) & = (\pi(s(0))-\mu) \\
\rho(s, p_{\pi(x) \leq \mu},t) & = (\mu - \pi(s(0))) \\
\rho(s, \neg \phi,t) & =-\rho(s, \phi, t) \\
\rho(s, \phi_1 \wedge \phi_2, t) & =\min \left\{\rho\left(s, \phi_1, t\right), \rho\left(s, \phi_2, t\right)\right\} \\
\rho(s, \phi_1 \vee \phi_2, t) & =\max \left\{\rho\left(s, \phi_1, t\right), \rho\left(s, \phi_2, t\right)\right\} \\
\rho(s, \phi_1 \mathcal{U}_{[a, b]} \phi_2, t) & =\max _{t_u \in[a, b]}\left\{\operatorname { m i n } \left\{\rho\left(s, \phi_2, t\right),\right.\right. \\
& \left.\left.\operatorname { m i n } _{t^{\prime} \in\left[0, t_u\right)}\left\{\rho\left(s, \phi_1, t\right)\right\}\right\}\right\} \\
\rho(s, F_{[a, b]} \phi, t) & =\max _{t_u \in[a, b]}\left\{\rho\left(s, \phi, t\right)\right\} \\
\rho(s, G_{[a, b]} \phi, t) & =\operatorname { m i n } _{t_u \in[a, b]}\left\{\rho\left(s, \phi, t\right)\right\}\:.
\end{array}
\end{equation}
The robustness degree is sound, i.e., $s\models\phi \iff \rho(s,\phi,t)\geq 0$.

\section{Problem Statement}

\subsection{Kinodynamic Constraints}
\label{sec:problem}
Let $S = \left(f, X, U, x_{init}\right)$ be a dynamical system, where $X \subseteq \mathbb{R}^n$ and $U \subseteq \mathbb{R}^m$ are the bounded state and control spaces. The state space $X$ contains obstacles $X_{obs} \subset X$ and the free space is denoted by $X_f = X$ $\symbol{92}$ $X_{obs}$.
$f$: $X \times U \rightarrow X$ is a Lipschitz continuous function, and $x_{\text{init}}$ is the initial state of the system. The system behavior is dictated by differential equations of the following form:
\begin{equation}
    \dot{x}=f(x(t), u(t))
\end{equation}
Where, $x(t) \in X , u(t) \in U$. we denote by $x[x_{init}, u]$ the state trajectory originating at $x_{\text{init}}$ obtained by implementing control policy $u$. Let $\upsilon = \{u: R_{\geq0} \rightarrow U\}$ be the set of all control policies. The system $S$ is said to satisfy an STL specification $\phi$ under a control policy $u \in \upsilon$ if the state trajectory starting at $x_0$ satisfies $\phi$, i.e., $x[x_\text{init}, u] \models \phi$.

\subsection{Mission Specification}
% \todo{we are agnostic as to how Psi is obtained. Thus, Psi is directly part of our problem formulation. We will emphasize this point more clearly in our revised draft}
In this work, we focus primarily on the development of a highly efficient trajectory planner to use in the context of a temporal logic planning problem. As such, we assume the existence of a \emph{candidate} solution, which can be obtained via solving a MILP using a low-fidelity motion model~\cite{belta2019formal,raman2014model}, applying SMT-based reasoning over an abstract syntax tree representation~\cite{leahy2022fast,cardona2023flexible}, or by using a logic such as Time Window Temporal Logic~\cite{TWTL}. Typical approaches to solving plan synthesis with a MILP use low-fidelity motion models to identify a discrete motion plan, which is then implemented by a continuous motion planner~\cite{leahy2021scalable}. 
For non-trivial motion models, a low-fidelity model may be highly inaccurate or produce infeasible trajectories. Thus, we consider an intermediate specification consisting of disjoint regions to be visited, each with its own time intervals. That is, we assume a high-level planner produces a specification $\Psi$ such that for a given signal $s$, $s\models\Psi\implies s\models \phi$. However, since the high-level planner does not account for kinodynamic motion plans, satisfaction of the (simpler) specification $\Psi$ may not be possible. 
Our approach is agnostic to how $\Psi$ is obtained. Thus, $\Psi$ is directly part of our problem formulation. Our goal is to find kinodynamic plans over a fragment of STL of the form:
%\begin{equation}\label{eq:fragment}
%\begin{aligned}
%    \psi = & \bigwedge_{i=1}^{n} \Big(\underbrace{\mathbf{F}_{[a_i,b_i]}(x \in X_{goal_i})}_{\text{time-bounded goals}} \land \underbrace{\mathbf{F}(x \in X_{goal_i})}_{\text{unbounded goals}}\Big)\:.
%\end{aligned}
%\end{equation}
\begin{equation}\label{eq:fragment}
\begin{aligned}
    & \Psi =  \left(\bigwedge_{i=1}^{n} \psi^i_{b}\right)\wedge\left( \bigwedge_{j=1}^{m} \psi^i_{un}\right)  \\
    & \psi^i_{b} =  \mathbf{F}_{[a_i,b_i]}(x \in X_{goal_i}),  
    & \psi^j_{un} =  \mathbf{F}_{[0,\infty)}(x \in X_{goal_j}) \\
\end{aligned}
\end{equation}
% \todo{Does “bounded” mean that the goal needs to be reached within a time-bound? It should be described to avoid misunderstanding from the readers.}
Where, $\psi_{b}, \psi_{un}$ represent bounded, and unbounded goals respectively. Unbounded goals are defined over the interval $[0,\infty)$.  

\textit{Problem 3.1:} Given a dynamical system $S$ and an STL specification $\Psi$ as written in \eqref{eq:fragment}, find a control policy $u$ such that the system satisfies $\Psi$ under policy $u$, and the cost function (based on the robustness metric) for the state trajectory is minimized.
It will be necessary to assume that the problem can be solved using trajectories generated by piecewise constant control functions. This is a reasonable way to generate a trajectory using a computational approach~\cite{SST}.

%\textcolor{red}{where should the assumptions be ?, directly cite lemma from SST paper?  The assumption on decompose specification so we can find geometric paths for the solution.}

\section{Proposed Approach}
In this section, we describe how the proposed algorithm finds a kinodynamically feasible path that satisfies the specification $\Psi$, as outlined in Algorithm~\ref{alg:high-level-planner}.
First, \textbf{CandidatePlans}$(\Psi)$ enumerates all possible orders in which the spatial regions can be visited while adhering to time constraints. Next, \textbf{GeoPlanner} generates a geometric path that visits these regions in the specified order. Finally, this geometric path is passed to our \textbf{LG-SST-STL} planner (modified \textbf{SST} planner \cite{SST}) which utilizes the geometric path as a guide for sampling, optimizing for the robustness value of the formula $\Psi$. If \textbf{LG-SST-STL} returns a positive robustness value, this indicates that the found kinodynamic path is valid, and the process concludes.
% \todo{Add SST Section}

%An STL specification is comprised of multiple constraints.  we will be using a decoupled approach where different parts of the STL formulation combination of biased-sampling, kinodyanmic planning, collision checking and cost-optimization to satisfy the STL specification.
%\todo{To discuss with Kevin}
%
%\begin{enumerate}
%    \item Satisfying the kinodynamic constraints equation $f(x(t),u(t))$. These constraints are satisfied by definition by using a forward propagation planner such as \sst  
%%    \item Satisfying global constraints of the form $G\_{[a, b]}(\phi)$. Since these constraints must always be true , e.g. such as obstacle avoidance, they can simply be satisfied by the collision checking.  If these contraints are time-bounded we can simply modify the collision checker by enforcing these constraints during that specific timestep, this is easy as the kinodynamic trajectory produced by \sst is directly time-parameterized.  
%    \item Satisfying spatial constraints. We assume a simple task-planner described in \label{subsec:task_planner}, which can give us a discrete order of spatial regions that need to be visited that if time parameterized can potentially satisfy the time constraints. Using this simple      
%    \item Satisfying the temporal constraints. The temporal constraints will be satisfied and optimized using a specific  cost function for RRT the using the  Having the discrete order of regions we need to      
%\end{enumerate}

\begin{algorithm}[h]
    \caption{High Level Planner}
    \label{alg:high-level-planner}
    \SetNoFillComment
    \KwInput{$S = (f, X, U, x_{init})$, $N_{max}$, $T_{max}$, $J(x,u)$, $\Psi$}
    \KwOutput{Optimal Control Policy $u^*$}

    $\{M_1, M_2, ..., M_m\} \leftarrow$ CandidatePlans($\Psi$) \\
    
    \For{$i \leftarrow 1$ \KwTo $m$}{
        $\{GP, L_{max}\} \leftarrow$ GeoPlanner($M_i$)\\
        $\{u, \rho \} \leftarrow$ LG-SST-STL($\{S, GP, L_{max}\}$) \\
        \If {$\rho \geq 0$}{
            \Return u
            }
    }
\end{algorithm}

\subsection{Discrete Region Orders}\label{subsec:task_planner}
\begin{algorithm}[h] \label{alg:plans}
    \caption{CandidatePlans}
    \label{alg:task-planner}
    \SetNoFillComment
    \KwInput{$\Psi$} 
    \KwOutput{$M_1, \ldots ,M_m$}
    $constraints \leftarrow  \{ \empty \}$ \\ \label{line:init-constraints}
    \For{$i = 1$ to $n$}{ \label{line:outer-loop}
        \For{$j = 1$ to $n$}{ \label{line:inner-loop}
            \If{no\_time\_overlap($\psi^i_b, \psi^j_b$)}{ \label{line:overlap-check}
                $constraints  \leftarrow \{constraints, (i,j)\} $ \\ \label{line:add-constraint}
            }
        }
    }
    $\mathcal{M} \leftarrow  \{ \empty \}$ \\ \label{line:init-M}
     \For {$M \in  perm(\psi_{b}, \psi_{un})$}{ \label{line:perm-loop}
            \If{$no\_violation(constraints, M )$}{ \label{line:violation-check}
             $\mathcal{M}  \leftarrow \{\mathcal{M}, M \} $ \\ \label{line:add-M}
            } 
    }
    \Return $\mathcal{M}$ \label{line:return}
\end{algorithm}
Algorithm \ref{alg:task-planner} comes up with all possible orders of regions that need to be visited sequentially to satisfy the formula $\Psi$. In \textbf{algorithm \ref{alg:task-planner}}, \textbf{lines~\ref{line:outer-loop}-\ref{line:add-constraint}}, all bounded goals, denoted by $\psi_b$, are evaluated for potential time overlaps. If there is no overlap, one region must be visited strictly before the other, and these orderings are added to the set of constraints. Subsequently, in \textbf{Algorithm \ref{alg:task-planner}}, \textbf{lines~\ref{line:perm-loop}-\ref{line:add-M}}, we generate all possible permutations of the regions to be visited and check for any violations of the ordering constraints derived from the time intervals. If no constraints are violated, the permutation is added to the set of candidate paths.

This algorithm has a worst-case time complexity of $\mathcal{O}(n!)$ in terms of the number of potential candidate paths to be planned by the motion planner. Nonetheless, as it enumerates all possible orders, it is complete. Additionally, for complex nonlinear dynamic models where computing steering functions are intractable, we argue that alternative methods encoding the full formula $\phi$ as a single cost function in an \textsc{ao} planner fail to achieve convergence to positive robustness, as demonstrated by our experimental results.

\subsection{Continuous Spatial Path}

The candidate sequence regions $M = \{X_{goal_1}, X_{goal_2}, \ldots, X_{goal_n}\}$, to be visited, serves as a series of intermediate goals, where $X_{goal_1}=x_{init}$ is the initial state and $X_{goal_n} = x_{goal}$ is the final goal. To construct the lead geometric path $GP$, we utilize a geometric planner, such as Rapidly-exploring Random Trees (\rrts). The path is generated as the sum of sub-paths, where each sub-path connects two consecutive intermediate goals, $\text{GeoPlan}(X_{goal_i}, X_{goal_{i+1}})$ for $i = 1, \ldots, n-1$, as described in \textbf{Algorithm~\ref{alg:decomp}}, \textbf{Line ~\ref{geo:geoplan}}.
\begin{algorithm}[h]
    \caption{GeoPlanner}
    \label{alg:decomp}
    \SetNoFillComment
    \KwInput{$M$} \label{geo:input}
    \KwOutput{$GP, L_{max}$} \label{geo:output}
    $\{X_{goal_1}, X_{goal_2}, \ldots, X_{goal_n}\} = M $ \\ \label{geo:subgoals}
    $X_{goal_1} = x_{init}$, $X_{goal_n} = x_{goal}$\\ \label{geo:init-goal}
    $GP \leftarrow \sum_{i=1}^{n-1} \text{RRT*}(X_{goal_i}, X_{goal_{i+1}})$ \\ \label{geo:geoplan}
    $L_{max} \leftarrow 2n -1 $ \label{geo:layers}
\end{algorithm}
We propose decomposing the global geometric path $GP$ into sub-regions called \textbf{Layers}. We construct one layer for each subpath $(X_{goal_i}, X_{goal_{i+1}})$ and for each subgoal $X_{goal_i}$ for a total $2n-1$ layers. An illustrative example is shown in \autoref{fig:layer-assign}, where intermediate goals and their connecting subpaths are structured into layers. These layers define regions that guide biased sampling, enforce the correct visitation order of the tree, and enable crossovers when necessary based on the sequence of regions.
%By incorporating this layered representation, supplementary information is encoded into each node of the planner's tree, enhancing its ability to integrate contextual data during execution (\textbf{Algorithm~\ref{alg:decomp}}). This approach forms a foundational element for incorporating both temporal and spatial constraints.
% generat goal each associated with a layer number. Each layer corresponds to the position of the sub-path in the sequence of intermediate goals $M$. The layering process assigns numbers to each sub-path and its associated intermediate goals, encoding hierarchical information that can be used during the motion planning process (\textbf{Algorithm~\ref{alg:decomp}}, \textbf{Line ~\ref{geo:layers}}). 
%The global geometric path $GP$ serves as guidance, narrowing the focus to regions critical for satisfying \stl constraints. Using the path provided by $GP$, we propose modifications to the Sparse Stable Trees (SST) planner. 
%The modified planner incorporates the layer numbers assigned to sub-paths and intermediate goals, enabling it to prioritize critical regions. This planner is referred to as the \textbf{Layer Guided SST with STL Cost (LG-SST-STL)}. The layered guidance and \stl-driven optimization contribute to improved efficiency and accuracy in dynamically complex environments.
\subsection{Kinodynamic Spatial Temporal Path}
In this section, we explain in detail the modifications we propose and outline how our framework integrates with the \sst \cite{SST} planner. Modifications are highlighted in \textbf{green} in \textbf{Algorithm~\ref{alg:LG-SST-STL}}, and the following subsections explain each modification in detail.

\begin{algorithm}
    \caption{LG-SST-STL}
    \label{alg:LG-SST-STL}
    \SetNoFillComment
    
    \KwInput{$S = (f, X, U, x_{init})$, $N_{max}$, $T_{max}$, $J(x,u)$, $GP, L_{max}$} \label{SST-STL:input}
    \KwOutput{Optimal Control Policy $u^*$} \label{SST-STL:output}

    $\mathcal{T} = \{x_{init}\}$; $\text{Cost}(x_{init}) = 0$ \\ \label{SST-STL:init}
    \For{$i = 1$ to $N_{max}$}{ \label{SST-STL:main-loop}
        \textcolor{darkgreen}{$L_{rand}, x_{rand} \leftarrow$ SpatialBiasSampler($X,L_{max}$)} \\ \label{SST-STL:sample-state}
        $x_{near} \leftarrow$ NearestNeighbor($\mathcal{T}$, $x_{rand}$) \\ \label{SST-STL:NN}
        $u_{rand} \leftarrow$ Sample($U$) \\ \label{SST-STL:sample-control}
        $T_{rand} \leftarrow$ Sample($0$, $T_{max}$) \\ \label{SST-STL:sample-time}
        $x[t] \leftarrow x_{near} + \int_{0}^{T_{rand}} f(x(\tau), u_{rand}) d\tau$ \\ \label{SST-STL:propagation}
        \If{$x[t] \in X_f$}{ \label{SST-STL:collision-check}
            \If{\textcolor{darkgreen}{dist($x[t]$ - $GP$) $\leq r_{prop}$}}{ \label{SST-STL:propagation-radius-check}
                \textcolor{darkgreen}{$L(x[t]) \leftarrow LayerAssign(x(t))$} \\ \label{SST-STL:layer-assign}
                \If{\textcolor{darkgreen}{$| L(x_{near}) - L(x[t]) | \leq 1$}}{ \label{SST-STL:layer-check}
                    $\text{newCost} \leftarrow \text{Cost}(x_{near}) + J(x[t], u_{rand})$ \\ \label{SST-STL:cost-assign}
                    \If{$\text{newCost} < \text{Cost}(x[t])$ \textbf{or} $x[t] \notin \mathcal{T}$}{
                        AddNode($\mathcal{T}$, $x[t]$) \\ \label{SST-STL:cost-check}
                        $\text{Cost}(x[t]) \leftarrow \text{newCost}$ \\ \label{SST-STL:cost-update}
                        UpdateTree($\mathcal{T}$, $x[t]$) \label{SST-STL:tree-update}
                    }
                }
            }
        }
        \If{IsGoalReached($x[t]$, $x_{goal}$)}{ \label{SST-STL:goal-check}
            \Return ReconstructPath($\mathcal{T}$, $x[t]$), $u[t]$ \label{SST-STL:get-trajectory}
        }
    }
    \Return Failure
\end{algorithm}
\noindent
% We introduce four major modifications \textit{\textbf{Biased Sampling strategy, Planner Layers, Propagation Radius, and modified Cost Function}} to the SST planner \cite{SST}.

\subsubsection{Biased Sampling}
We use the global geometric path $GP$ to bias the sampling strategy. This strategy confines sampling to regions proximal to the global path. To control this proximity, we introduce a new parameter called the \textbf{sampler selection radius ($s_r$)}, which restricts sampling to a neighborhood around $GP$. The sampling process is described in \textbf{Algorithm~\ref{alg:sampler}}, specifically \textbf{Line~\ref{sampler:state}}, and is illustrated in Figure~\ref{fig:layer-assign} which is the outer layer shaded in the color orange. By focusing sampling efforts along the guidance path, the planner achieves higher computational efficiency. This is because the geometric path provides a partial solution for the spatial component of the STL specification, significantly reducing the search space.

\begin{algorithm}[h]
    \caption{Spatial Bias Sampler}
    \label{alg:sampler}
    \SetNoFillComment
    \KwInput{$X, L_{max}$} \label{sampler:input}
    \KwOutput{$x_{rand},L_{rand}$} \label{sampler:output}
    $L_{rand} \leftarrow$ Sample($0,L_{max}$)\\ \label{sampler:layer}
    $x_{rand} \leftarrow$ BiasSampler($X,s_r$) \label{sampler:state}
\end{algorithm}
% \begin{figure}[htbp]
%     \centering
%     \includegraphics[width=1\linewidth]{imgs/3_Sampler.png}
%     \caption{Bias Sampling Region}
%     \label{fig:gp-sampler}
% \end{figure}

\subsubsection{Layer Assignment}
\begin{figure}[htbp]
    \centering
    \includegraphics[width=0.8\linewidth]{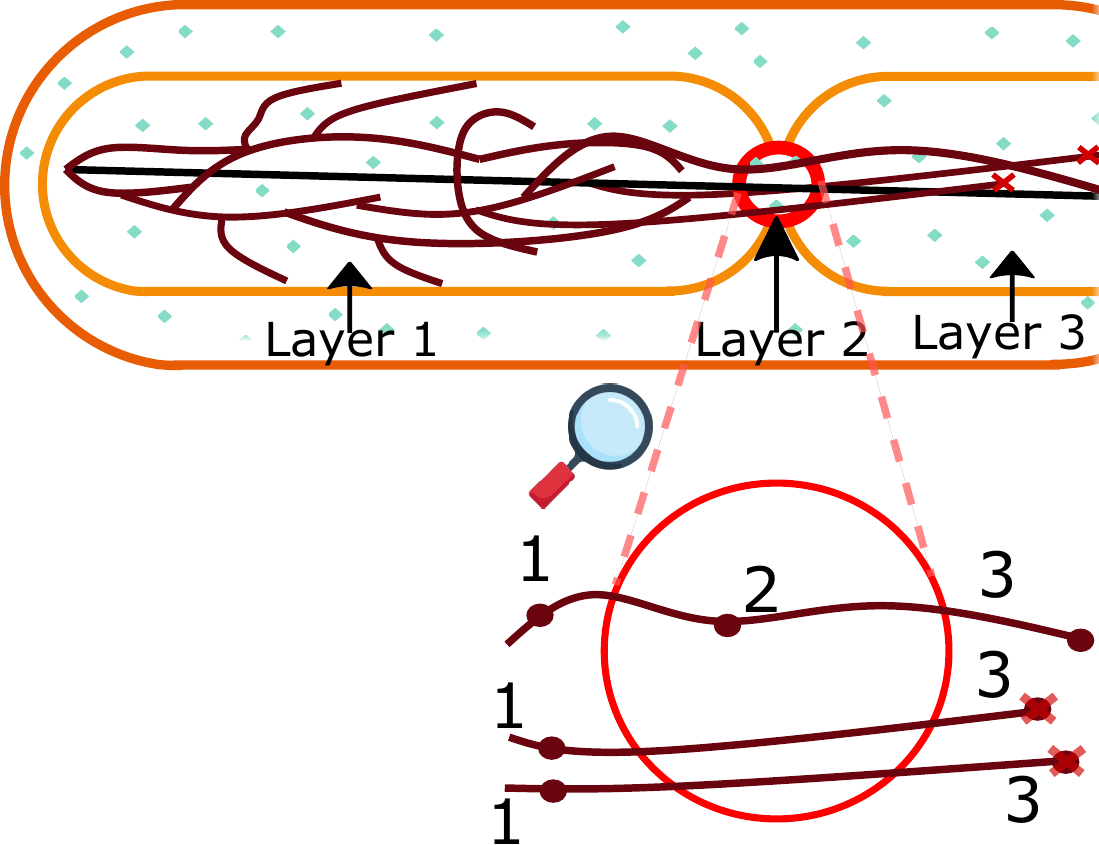}
    \caption{Layer Assignment and connection of nodes, the magnified image shows the node that are rejected.}
    \label{fig:layer-assign}
\end{figure}

The layers assigned to the global path earlier are now used to guide the planner during sampling and tree growth. Specifically, each sampling iteration selects a layer, $L_{rand}$, and a sample, $x_{rand}$, within that layer as shown in \textbf{Algorithm~\ref{alg:sampler}}, \textbf{Line~\ref{sampler:layer}}. This ensures that the planner grows the tree within the selected layer, as illustrated in Figure~\ref{fig:layer-assign}. Layers are also assigned to each node in the planner’s tree based on proximity to the corresponding layer of $GP$. This assignment is described in \textbf{Algorithm~\ref{alg:layer-check}}, \textbf{Line~\ref{layer:assign}}. A new node, $x[t]$, is only added to the tree if it belongs to a layer consecutive to its parent node, as enforced in \textbf{Algorithm~\ref{alg:LG-SST-STL}}, \textbf{Line~\ref{SST-STL:layer-check}}. This restriction ensures that the planner maintains consistency with the \textsc{stl} specification and avoids undesired behavior by connecting with incorrect nodes in the tree.
% The layers that were assigned to the global path earlier will now be used to select where we sample a point for the planner for each iteration. This process is shown in detail in algorithm \ref{alg:sampler}, where we randomly select a layer and then randomly select a sample in that layer (\textbf{Algorithm~\ref{alg:sampler}}, \textbf{Line ~\ref{sampler:layer}}), ensuring tree growth in each layer. Additionally, we can introduce heuristics to further speed up the process, given prior information on the problem.
% Layers will also be assigned to each node in the tree, corresponding to the closest layer in the global path as described in \textbf{algorithm \ref{alg:layer-check}} as shown in figure \ref{fig:layer-assign}. Assigning layers to nodes introduces a new dimension, which we leverage to modify how the tree grows. Once the layer is assigned to a new node in the tree, we only allow it if its parent node is in a consecutive layer (\textbf{Algorithm~\ref{alg:LG-SST-STL}}, \textbf{Line ~\ref{SST-STL:layer-check}}); this helps ensure that no component of specification is avoided by the planner, as shown in figure \ref{fig:layer-assign}. This also helps us solve problems with self-intersection paths (i.e. loops).
% \todo{Improve explanation of figure and minor technical details }
\begin{algorithm}[h]
    \caption{Layer Assignment}
    \label{alg:layer-check}
    \SetNoFillComment
    \KwInput{$x[t]$} \label{layer:input}
    \KwOutput{$L(x[t])$} \label{layer:output}
    $L(x[t]) \leftarrow \arg\min_{l \in \{1,\ldots,L_{max}\}} \text{dist}(x[t], GP_l)$ \label{layer:assign}
\end{algorithm}
\subsubsection{Propagation Radius}
\begin{figure}[h]
    \centering
    \includegraphics[width=0.8\linewidth]{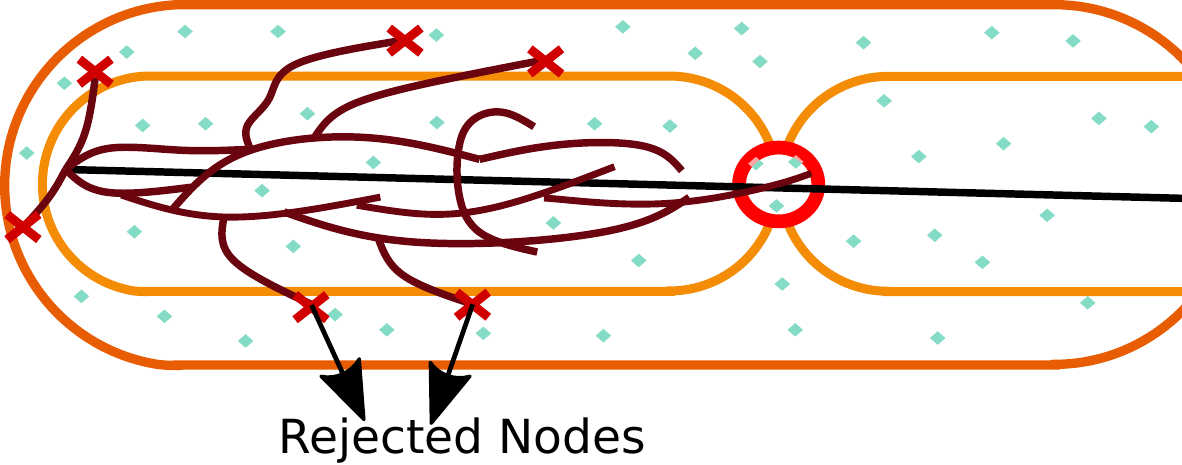}
    \caption{The figure shows rejection of nodes that deviate more than allowed by the propagation radius}
    \label{fig:prop-radius}
\end{figure}

We introduce a new parameter, the \textbf{propagation allowed radius ($r_{prop}$)}, which limits the planner’s growth to remain within a defined radius t the global path $GP$. This restriction is enforced in \textbf{Algorithm~\ref{alg:LG-SST-STL}}, \textbf{Line~\ref{SST-STL:propagation-radius-check}}. If a propagated node $x[t]$ exceeds $r_{prop}$, it is rejected, as shown in Figure~\ref{fig:prop-radius}.
By limiting deviations from the global path, the propagation radius ensures that the planner remains focused on regions critical for satisfying \textsc{stl} constraints. Adjusting $r_{prop}$ can help balance between exploration and adherence to the guidance path.
\subsubsection{Satisfy the Temporal constraints (Cost Function)} 
The above methodology will help us create a kinodynamic path that will satisfy all the spatial components of the \textsc{stl} formulae by design, and it will additionally be time-parameterized. 
Cost functions have been used before in many different scenarios, such as \cite{RRT*-STL, RT-RRT*-STL, scLTL-RRT*}. We will be using a simple cost inspired by \cite{RT-RRT*-STL} that will ensure that as it decreases, the \textsc{stl} specification will be satisfied. We define a simple cost function $J$ that includes the robustness of an \textsc{stl} specification. For calculating cost after each new node that is added to the tree $\mathcal{T}$, we make the cost compute recursive, which helps reduce overhead. 
\begin{align}
\bar{\rho}(\mathbf{x}_i, (f(\xi_i) \sim \mu), t) &= \begin{cases}
    \mu - f(\xi_i(t)) & \sim=\leq \\
    f(\xi_i(t)) - \mu & \sim=\geq
\end{cases} \\[1em]
\bar{\rho}(\mathbf{x}_i, \phi_1 \wedge \phi_2, t) &= \begin{cases}
    \star & \text{if } \bar{\rho}(\mathbf{x}_i, \phi_1, t) \text{ and} \\
    & \quad \bar{\rho}(\mathbf{x}_i, \phi_2, t) = \star \\
    \bar{\rho}(\mathbf{x}_i, \phi_1, t) & \text{if } \bar{\rho}(\mathbf{x}_i, \phi_2, t) = \star \\
    \bar{\rho}(\mathbf{x}_i, \phi_2, t) & \text{if } \bar{\rho}(\mathbf{x}_i, \phi_1, t) = \star \\
    \min(\bar{\rho}(\mathbf{x}_i, \phi_1, t), & \quad \bar{\rho}(\mathbf{x}_i, \phi_2, t))
\end{cases} \\[1em]
\bar{\rho}(\mathbf{x}_i, F_{[a,b]}\phi, t) &= \begin{cases}
    \star & \text{if } t < a \text{ or } t > b \\
    \bar{\rho}(\mathbf{x}_i, \phi, t) & \text{if } t = a \\
    \max(\bar{\rho}(\mathbf{x}_i, \phi, t), & \quad \bar{\rho}_{\text{parent}}(\mathbf{x}_i, F_{[a,b]}\phi))
\end{cases} \\[1em]
\bar{\rho}(\mathbf{x}_i, F\phi, t) &= \max(\bar{\rho}(\mathbf{x}_i, \phi, t), \bar{\rho}_{\text{parent}}(\mathbf{x}_i, F\phi))
\end{align}
Hereinafter, we define the robustness value $\bar{\rho}$ associated to a node $\mathbf{x}_i$ in $\mathcal{T}$. Since a trajectory until a given node in the LG-SST-STL tree might be partially testable against the STL specification, we recursively define the robustness of (partial) trajectories until node $\mathbf{x}_i$. Consider a simple specification like $F_{[4,5]}(x>2)$. It is not possible to assess the trajectory's robustness against the specification until the first observation of the trajectory for time $t=4$ has been made. As highlighted in \cite{onlinemonitoringsignal}, we use and maintain a syntax tree of STL formula in memory, augmented with a robustness value $\bar{\rho}$ associated with the nodes in $\mathcal{T}$. Temporal operators are equipped with robustness values $\bar{\rho}$ for nodes $\mathbf{x}_i \in \mathcal{X}$, that are stored. Further, we define by $\bar{\rho}_{\text {parent }}: \mathcal{X} \times \Phi^{\Sigma} \rightarrow \mathbb{R} \cup\{\star\}$ the $\bar{\rho}$ value of a temporal operator in the syntax tree of the specification, for the parent of node $\mathbf{x}_i$ in the LG-SST-STL tree, where $\star$ is a dummy symbol used to provide no real value to a formula, the robustness of which cannot be stated for a given trajectory (i.e., for trajectories shorter than the lower bound of the interval of a temporal operator). In the following, $\bar{\rho}_{\text {parent }}$ is called to compute the actual value of $\bar{\rho}$ for node $\mathbf{x}_i$ (since $\bar{\rho}$ depends on the value of $\bar{\rho}$ for the parent node of $\mathbf{x}_i$ ). The robustness value $\bar{\rho}: \mathcal{X} \times \Phi^{\Sigma} \times \mathbb{N} \rightarrow \mathbb{R} \cup\{\star\}$ associated to a node $\mathbf{x}_i$ in $\mathcal{T}$ is given by the recursive function:

Timed temporal operators, the design of $\bar{\rho}$ only uses the evaluation of predicates for the relevant time intervals. Outside of these, predicates are not evaluated, hence are not reflected in the calculation of the cost function. Also, the definition of $\bar{\rho}$ as such enables an easy computation of the $\min$ and $\max$ in the case of temporal operators: for a given node $\mathbf{x}_i$, the whole trajectory until node $\mathbf{x}_i$ doesn't need to be tested, but only the spatial coordinates of node $\mathbf{x}_i$ and the value of $\bar{\rho}$ of the temporal operator for the parent of $\mathbf{x}_i$, which leads to lower computational complexity. We now define $\bar{\bar{\rho}}: \mathcal{X} \times \Phi^{\Sigma} \times \mathbb{N} \rightarrow \mathbb{R}$, that will be directly called by the LG-SST-STL cost function:$J\left(\mathbf{x}_i\right)=J_\phi\left(\mathbf{x}_i\right) = \bar{\bar{\rho}}\left(\mathbf{x}_i, \phi, t\right)= -\min \left(\bar{\rho}\left(\mathbf{x}_i, \phi, t\right), 0\right)$

% The advantage of this cost is that it can be evaluated over partial trajectories until node \textbf{x\_i}. Now, we need to give an example similar to the Jumova paper to explain this. But the main idea is that we will be using the $\ast$ symbol for areas that we are not computing, e.g., and are outside the time limits.% \todo{Add figure to explain the cost here. We will adapt it for our own purposes}

\begin{figure*}[t]
    \centering
    \includegraphics[width=0.85\textwidth]{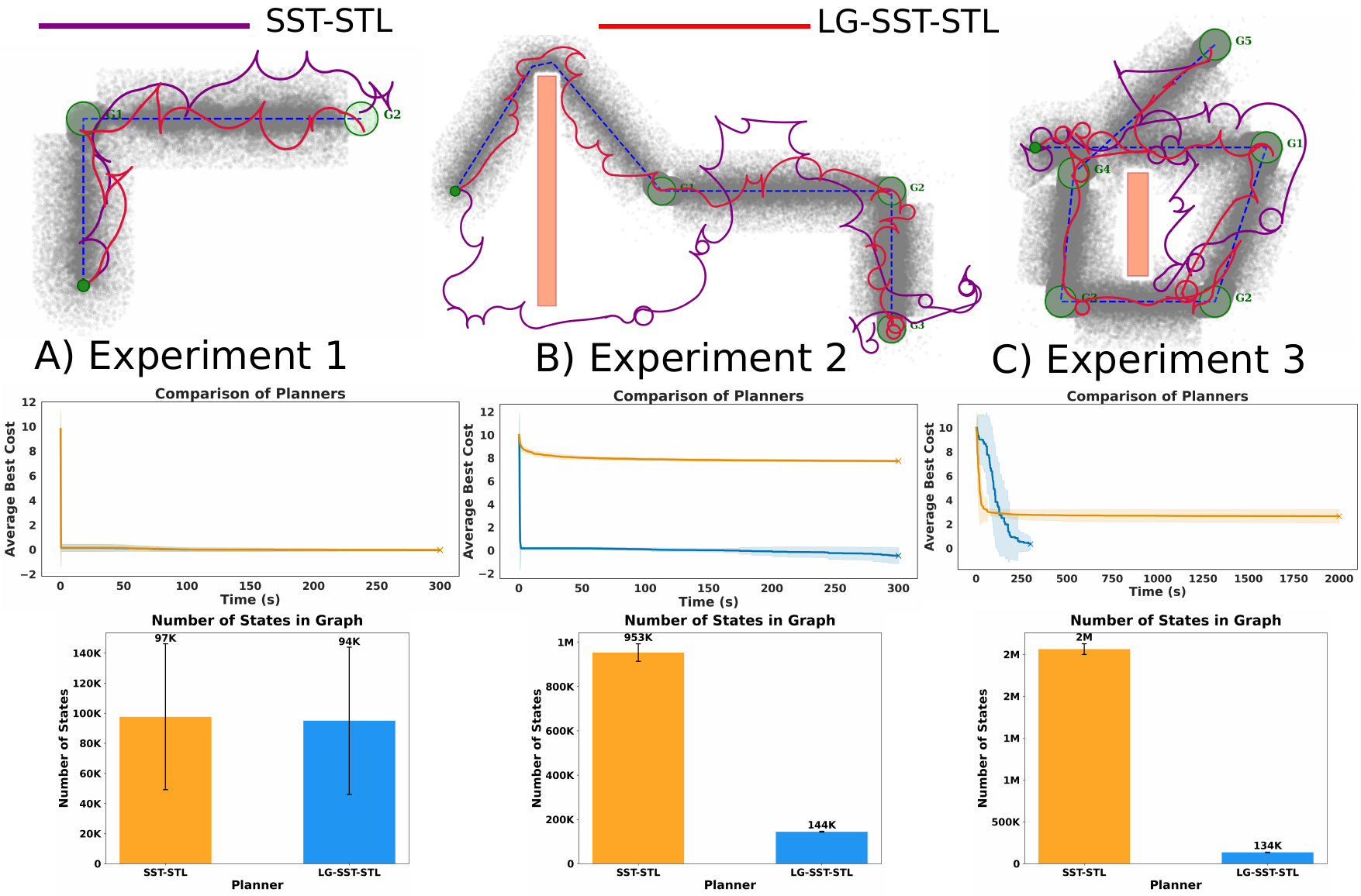}
    \caption{Comparison of the SST-STL and LG-SST-STL planners across three experiments.
    The middle row displays the average best cost achieved by each planner over time. The bottom row compares the number of states in the graph.}
    \label{fig:experiments}
\end{figure*}

\section{Case Studies}
We validate our approach through simulations conducted using the Open Motion Planning Library (OMPL)\cite{sucan2012the-open-motion-planning-library}, for our implementation \cite{LG-SST-STL} . Our experiments encompass three environments, each designed to test an Ackermann-steered vehicle navigating in $SE(2)$ configuration space, where the state vector $\mathbf{x} = [x, y, \theta]^T \in \mathbb{R}^2 \times \mathbb{S}^1$ represents the vehicle's position and orientation. The system dynamics follow the standard Ackermann-steering model: $\dot{x} = v \cos(\theta), \quad \dot{y} = v \sin(\theta), \quad \dot{\theta} = \frac{v}{L}\tan(\delta)$
where $v$ is the linear velocity, $L$ is the wheelbase length, and $\delta \in [-\delta_{max}, \delta_{max}]$ is the steering angle. Each environment presents unique challenges and incorporates different \textsc{stl} specifications. For the experiments, the intermediary goal threshold $\varepsilon$ is set to $0.3\,\text{m}$, ensuring the vehicle comes sufficiently close to each goal region. We evaluate our approach against the baseline SST planner with STL cost using OMPL's benchmarking tools~\cite{moll2015benchmarking-motion-planning-algorithms}. The experimental evaluation consists of 60 runs for each planner across three distinct environments. For each run, we analyze the evolution of the \textsc{STL} cost $J$ over a time horizon of 300 seconds, along with the computational resources required by each planner. The comparative results of both performance metrics are presented in Figure~\ref{fig:experiments}.

\textbf{Experiment 1:}
The Objective of Experiment 1 is to evaluate the capability of the SST-STL planner and the proposed LG-SST-STL planner in sequentially achieving two unbounded spatial goals without incorporating time bounds into the cost function. This experiment assesses the planners' ability to generate feasible paths that satisfy the given \textsc{STL} specifications.
$$
\begin{aligned}
   \Psi =  \mathbf{F}\big(d((x,y)-(5,4))\leq \varepsilon \big) \land \mathbf{F}\big(d((x,y)-(10,4)) \leq \varepsilon\big)
\end{aligned}
$$
As depicted in Figure \ref{fig:experiments}, both the SST-STL planner and the LG-SST-STL planner successfully converged to feasible solutions, generating paths that satisfy the specified STL fragment. 

% This experiment evaluates the planner's behavior with a combination of temporally bounded and unbounded goals, demonstrating the importance of sequential goal satisfaction even when direct paths to time-bounded goals exist. The STL specification for this experiment is formulated as:
% \begin{figure}
%     \centering
%     \includegraphics[width=\linewidth]{imgs/Exp2.png}
%     \caption{Enter Caption}
%     \label{fig:enter-label}
% \end{figure}

% \begin{equation}
% \begin{aligned}
%    \varphi = & \mathbf{F}\big(|x-5| \leq \varepsilon \land |y-4| \leq \varepsilon\big) \land \\
%              & \mathbf{F}_{[35,65]}\big(|x-10| \leq \varepsilon \land |y-4| \leq \varepsilon\big)
% \end{aligned}
% \end{equation}
% A feature demonstrated in this experiment is the planner's adherence to the sequential ordering of goals. While a direct trajectory from start to goal 2 might appear more time-efficient, our planner successfully maintains the specified sequence by first visiting goal 1 (unbounded) before reaching goal 2 within its temporal constraints. This behavior showcases our method's capability to balance both sequential ordering requirements and temporal constraints, even when potentially shorter alternatives exist.

\textbf{Experiment 2:} consists of four sequential, time-bounded goals designed to evaluate the planner's ability to handle temporal constraints in navigation in presence of obstacle 
% measuring 0.4m × 2m, located at (2.3m, 1.5m). 
% The Signal Temporal Logic (STL) specification for this experiment is defined as follows:
% \begin{equation}
% \begin{aligned}
%    \Psi = & \mathbf{F}_{[0,3]}\big(|x-0.5| \leq \varepsilon \land |y-4| \leq \varepsilon\big) \land \\
%             &  \mathbf{F}_{[6,20]}\big(|x-5| \leq \varepsilon \land |y-4| \leq \varepsilon\big) \land \\
%              & \mathbf{F}_{[20,40]}\big(|x-10| \leq \varepsilon \land |y-4| \leq \varepsilon\big) \land \\
%             &  \mathbf{F}_{[35,65]}\big(|x-10| \leq \varepsilon \land |y-1| \leq \varepsilon\big),
% \end{aligned}
% \end{equation}
$$
\begin{aligned}
   \Psi = & \mathbf{F}_{[0,3]}\big(d((x,y) - (0.5,4)) \leq \varepsilon \big) \land \\
             & \mathbf{F}_{[6,20]}\big(d((x,y) - (5,4)) \leq \varepsilon\big) \land \\
             & \mathbf{F}_{[20,40]}\big(d((x,y) - (10,4)) \leq \varepsilon\big) \land \\
             & \mathbf{F}_{[35,65]}\big(d((x,y)- (10,1)) \leq \varepsilon\big),
\end{aligned}
$$
% where \( \mathbf{F}_{[t_1, t_2]} \) specifies that the corresponding spatial constraints must be satisfied within the time interval \([t_1, t_2]\).

As shown in \ref{fig:experiments}B, the \textsc{sst} planner with \textsc{stl} cost fails to find a path that meets all requirements. The planner opts for a longer path around the obstacle. This inefficiency prevents the planner from reaching some goals within respective time bounds, thus only partial satisfying STL. In contrast, the \textsc{LG-SST-STL} planner demonstrates significant improvements. By selecting the shortest route to Goal 1, it satisfies the time constraint and sequentially meets the deadlines for Goals 2 and 3. Notably, this approach requires ten times fewer graph states than the \textsc{sst} baseline.

\textbf{Experiment 3:} In this experiment, we demonstrate the planner's ability to handle crossovers. This capability is achieved through our layer based strategy, which restricts node connections to adjacent layers, effectively addressing the challenges posed by loops in path planning. The environment includes multiple goals with overlapping time bounds, creating a scenario where the order of visiting certain goals does not impact the overall satisfaction of the temporal logic specification. 
% Where \( \mathbf{F}_{[t_1, t_2]} \) specifies that the corresponding spatial constraints must be satisfied within the time interval \([t_1, t_2]\). The \textsc{stl} specification for this experiment is formulated as follows:
Due to the overlapping time bounds of multiple goals, the order in which these goals are visited is interchangeable, provided that each is reached within its time bounds. This results in multiple feasible sequences, and the planner must consider all permutations to find an optimal path. 
% \begin{equation}
% \begin{aligned}
%    \Psi = & \mathbf{F}_{[0,2]}\big(|x-0.5| \leq \varepsilon \land |y-4| \leq \varepsilon\big) \land \\ 
%              &  \mathbf{F}_{[6,20]}\big(|x-5| \leq \varepsilon \land |y-4| \leq \varepsilon\big) \land \\
%              & \mathbf{F}_{[20,30]}\big(|x-4| \leq \varepsilon \land |y-1| \leq \varepsilon\big) \land \\ 
%              & \mathbf{F}_{[25,90]}\big(|x-1| \leq \varepsilon \land |y-1| \leq \varepsilon\big) \land \\
%              & \mathbf{F}_{[30,120]}\big(|x-1.25| \leq \varepsilon \land |y-3.5| \leq \varepsilon\big) \land \\ 
%               &\mathbf{F}_{[35,150]}\big(|x-4| \leq \varepsilon \land |y-6| \leq \varepsilon\big),
% \end{aligned}
% \end{equation}
$$
\begin{aligned}
   \Psi = & \mathbf{F}_{[0,2]}\big(d((x,y)-(0.5,4)) \leq \varepsilon \big) \land \\
             & \mathbf{F}_{[6,20]}\big(d((x,y)-(5,4)) \leq \varepsilon\big) \land \\
             & \mathbf{F}_{[20,30]}\big(d((x,y)-(4,1)) \leq \varepsilon\big) \land \\
             & \mathbf{F}_{[25,90]}\big(d((x,y)-(1,1)) \leq \varepsilon \big) \land \\
             & \mathbf{F}_{[30,120]}\big(d((x,y)-(1.25,3.5)) \leq \varepsilon\big) \land \\
            &\mathbf{F}_{[35,150]}\big(d((x,y)-(4,6)) \leq \varepsilon\big),
\end{aligned}
$$

Even though our method needs to be run for each of $3! = 6$ permutations, it still requires less time and converges faster than the baseline planner. Our layer-guided SST planner efficiently navigates through these permutations by leveraging the geometric path, which reduces the search space and computational complexity. 
% As illustrated in Figure~\ref{fig:experiments}, our planner successfully finds a solution within 300 seconds. In contrast, the baseline SST-STL planner, even after running for seven times longer (2100 seconds), fails to converge to a solution, even though it has almost 14 times the amount of states in its graph.

% Include your figure here if needed
% \begin{figure}
%     \centering
%     \includegraphics[width=1\linewidth]{imgs/Exp3.png}
%     \caption{Comparison of the planning results for Experiment 3. The layer-guided SST planner successfully finds a solution, while the baseline SST-STL planner fails to converge.}
%     \label{Fig}C
% \end{figure}

% //TODO  : TALK ABOUT THE EXPERIMENTS AND ADD THE FORMULAS FOR EACH OF THE EXPERIMENTS//
%Experiment 1 : 4 Goals (4x Timed Eventually Timed  in a Conjunction)
%Experiment 2 : 3 Goals (Here we have an Untimed Goal in the middle and there is a Timed Goal at the End, L shape situation, we also tell as this is ordered, it chooses the right one and does not go from start to the last goal.
%Experiment 3 : Loops. 

% Define the system we use to run the experiments, including specs, etc.
% We want to show, \\
% 1. Loops \\
% 2. Different scenario examples \\
% 3. Compare - baseline, our approach with two different heuristics, etc. \\
% 4. Different Systems: Carlike/Differential-Drive/Holonomic Base ?
\section{Conclusion}
We developed a new approach to motion planning that efficiently handles spatiotemporal constraints. Our experiments demonstrated that our method significantly reduces computation time while handling complex scenarios with time-bounded goals. Notably the proposed method, can even produce crossover paths, if required by the specification. Future work will explore more expressive STL constraints, dynamic obstacles, and multi-robot systems.
\begin{acks}
The authors thank \textbf{Michael Da Silva} for his help in creating the diagrams that appear in this manuscript.
\end{acks}
%% The next two lines define the bibliography style to be used and
%% the bibliography file.

% \bibliographystyle{unsrt}
% \bibliography{sample-base}

\bibliographystyle{ACM-Reference-Format}
\bibliography{sample-base}

%%
%% If your work has an appendix, this is the place to put it.
% \appendix

% \section{Research Methods}

% \subsection{Part One}

\end{document}